\documentclass[runningheads]{llncs}
\usepackage[T1]{fontenc}

\usepackage{graphicx}

\usepackage{hyperref}

\usepackage{booktabs}
\usepackage[table,xcdraw]{xcolor}
\usepackage{multirow}
\usepackage{listings}
\usepackage{tikz}
\usepackage{orcidlink}
\usepackage{listings}
\usepackage{subcaption}
\usetikzlibrary{arrows.meta, positioning, shapes.geometric, shapes.symbols, shadows, backgrounds}

\definecolor{agentcolor}{RGB}{232, 245, 233} 
\definecolor{datacolor}{RGB}{255,243,224} 
\definecolor{boxcolor}{RGB}{245, 247, 250} 
\definecolor{bordercolor}{RGB}{52, 73, 94} 
\definecolor{highlight}{RGB}{41, 128, 185} 

\definecolor{darkgreen}{cmyk}{1, 0, 1, 0.5}

\begin{document}
%
%
\title{PMAx: An Agentic Framework for AI-Driven Process Mining\thanks{Preprint version submitted to EMMSAD 2026 (Tool Demonstration). This version has not undergone peer review.}}

%
\author{Anton Antonov\inst{1,2}\orcidID{0009-0004-1044-4884} \and Humam Kourani\inst{1,2}\orcidID{0000-0003-2375-2152}
\and Alessandro Berti\inst{2}\orcidID{0000-0002-3279-4795} \and Gyunam Park\inst{1}\orcidID{0000-0001-9394-6513}\and Wil M.P. van der Aalst\inst{2,1}\orcidID{0000-0002-0955-6940}
}

\authorrunning{A. Antonov et al.}

\institute{
Fraunhofer Institute for Applied Information Technology FIT, Schloss Birlinghoven, 53757 Sankt Augustin, Germany
\email{\{anton.antonov,humam.kourani,gyunam.park\}@fit.fraunhofer.de}
\and
RWTH Aachen University, Ahornstraße 55, 52074 Aachen, Germany
\email{\{a.berti,wvdaalst\}@pads.rwth-aachen.de}
}
\maketitle    
\begin{abstract}
Process mining provides powerful insights into organizational workflows, but extracting these insights typically requires expertise in specialized query languages and data science tools. Large Language Models (LLMs) offer the potential to democratize process mining by enabling business users to interact with process data through natural language. However, using LLMs as direct analytical engines over raw event logs introduces fundamental challenges: LLMs struggle with deterministic reasoning and may hallucinate metrics, while sending large, sensitive logs to external AI services raises serious data-privacy concerns. To address these limitations, we present PMAx, an autonomous agentic framework that functions as a virtual process analyst. Rather than relying on LLMs to generate process models or compute analytical results, PMAx employs a privacy-preserving multi-agent architecture. An Engineer agent analyzes event-log metadata and autonomously generates local scripts to run established process mining algorithms, compute exact metrics, and produce artifacts such as process models, summary tables, and visualizations. An Analyst agent then interprets these insights and artifacts to compile comprehensive reports. By separating computation from interpretation and executing analysis locally, PMAx ensures mathematical accuracy and data privacy while enabling non-technical users to transform high-level business questions into reliable process insights.


\end{abstract}
\section{Introduction}
Process Mining \cite{DBLP:books/sp/Aalst16} bridges the gap between model-based process management and data-oriented analysis. While the extraction of digital footprints from event logs provides immense operational transparency, conducting exploratory process mining, such as discovering complex process models, identifying bottlenecks, and checking conformance, remains a significant hurdle for non-technical domain experts. These tasks traditionally demand a high degree of proficiency in data manipulation and specialized mining frameworks like PM4Py \cite{DBLP:journals/simpa/BertiZS23} or PQL \cite{DBLP:books/sp/22/0001ABSGK22}. 

Recent advancements in Artificial Intelligence (AI), particularly Large Language Models (LLMs), have sparked strong interest in democratizing process mining. Recognizing this potential, major commercial vendors (e.g., Celonis\footnote{\url{https://www.celonis.com}.}, SAP Signavio\footnote{\url{https://www.signavio.com}.}, and Apromore\footnote{\url{https://apromore.com}.
}) have recently introduced AI copilots into their platforms, allowing users to query processes via natural language. While these commercial solutions effectively lower the barrier to entry for business users, a research gap exists for open, transparent, and locally deployable frameworks. For academic research and custom analytical pipelines, there is a distinct need for systems where the underlying analytical logic is inspectable, data handling is kept within the local environment, and the architecture is extensible.



Developing an autonomous AI system for process mining presents several technical and operational challenges. First, when prompted to directly compute throughput times or identify process variants, LLMs may hallucinate metrics or fabricate algorithmic outcomes \cite{DBLP:journals/pacmse/ZhangWWSMZCMZ25,DBLP:journals/corr/abs-2404-06035} due to their probabilistic nature. Second, real-world event logs are massive and typically contain highly sensitive organizational data. Transmitting this raw information to external APIs not only risks exceeding context window limits but also violates strict corporate data governance policies. To harness conversational AI without compromising accuracy or confidentiality, the system must shift from guessing to deterministic computation. This requires a paradigm where the agent autonomously synthesizes executable code to analyze the data \cite{DBLP:journals/sncs/BistarelliFMM25}. However, such an approach introduces a new vulnerability, as executing unverified generated scripts poses severe security risks \cite{DBLP:conf/uss/SpracklenWSMV25}. 

\begin{figure}[!t]
  \centering
  \includegraphics[width=\textwidth]{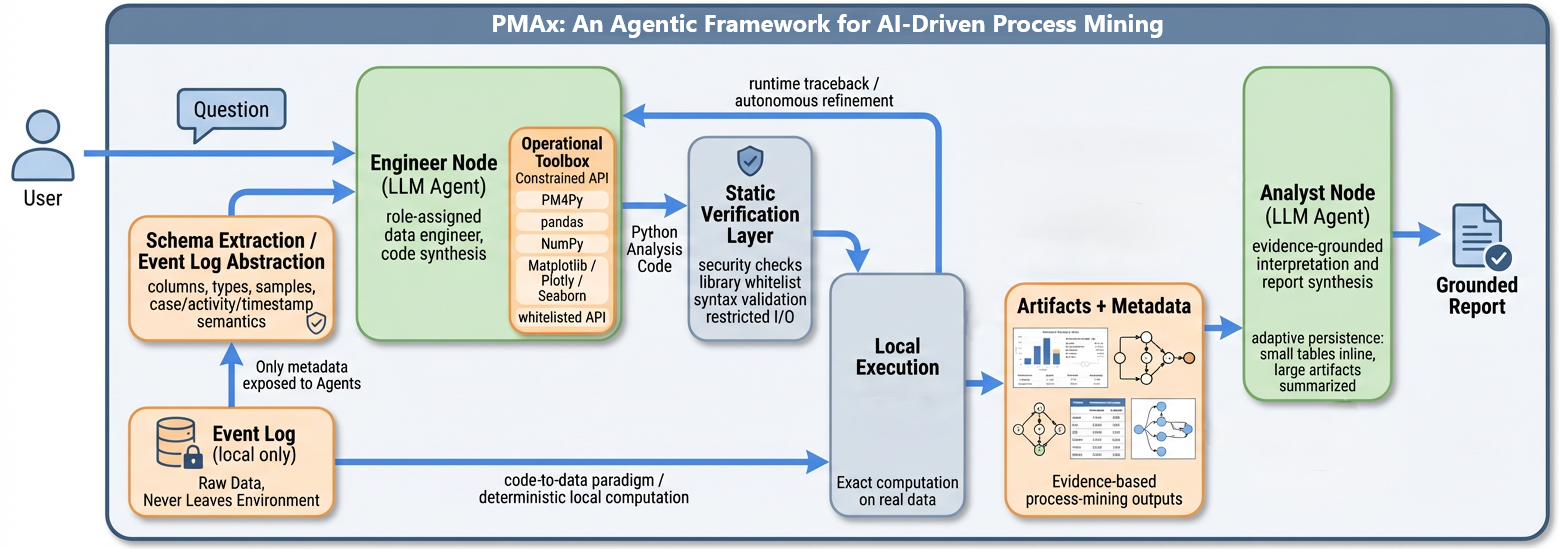}
  \caption{Overview of the proposed architecture.}
  \label{fig:overview}
\end{figure}

To this end, we propose \textbf{PMAx}\footnote{\url{https://github.com/fit-process-mining/ProMoAI}} (illustrated in \autoref{fig:overview}), an autonomous, agentic framework that orchestrates a secure, end-to-end process mining workflow. Our contribution is an orchestration framework defined by four pillars:
\begin{enumerate}
    \item \textbf{Secure Data Handling:} The framework extracts only lightweight structural metadata to provide the AI with context. The raw event data never leaves the user's local environment since only a static ``snapshot'' of the data in form of present columns with their corresponding type is provided. 

    \item \textbf{Process Mining Multi-Agent Workflow:} We employ a ``divide-and-conquer'' architecture to shift from probabilistic guessing to deterministic computation. An \textit{Engineer Agent}, grounded with specific process mining domain knowledge and library APIs, synthesizes executable code to create process mining artifacts (e.g., process models, summary tables, statistical charts). An \textit{Analyst Agent} then interprets these artifacts to compile comprehensive reports that combine textual insights with rich visual evidence. 
    
    
    \item \textbf{Reliable Execution:} The framework employs a controlled environment with a static verification layer to ensure system integrity. Within this environment, an autonomous self-correction loop captures runtime errors and automatically feeds them back to the agents for iterative refinement, ensuring reliability without human intervention.
    
    \item \textbf{Open-Source and Extensible Design:} Built on a standard Python ecosystem using established libraries like \cite{DBLP:journals/simpa/BertiZS23} and Pandas \cite{reback2020pandas}, PMAx is fully open-source. This provides a transparent and modular platform that can be easily deployed locally. Its architecture is designed for extensibility, allowing for integrating new process mining tasks or custom experimental pipelines without being tied to a proprietary ecosystem.

\end{enumerate}

\begin{figure}[!t]
\centering

\begin{subfigure}[t]{0.48\linewidth}
    \centering
    \includegraphics[width=\linewidth]{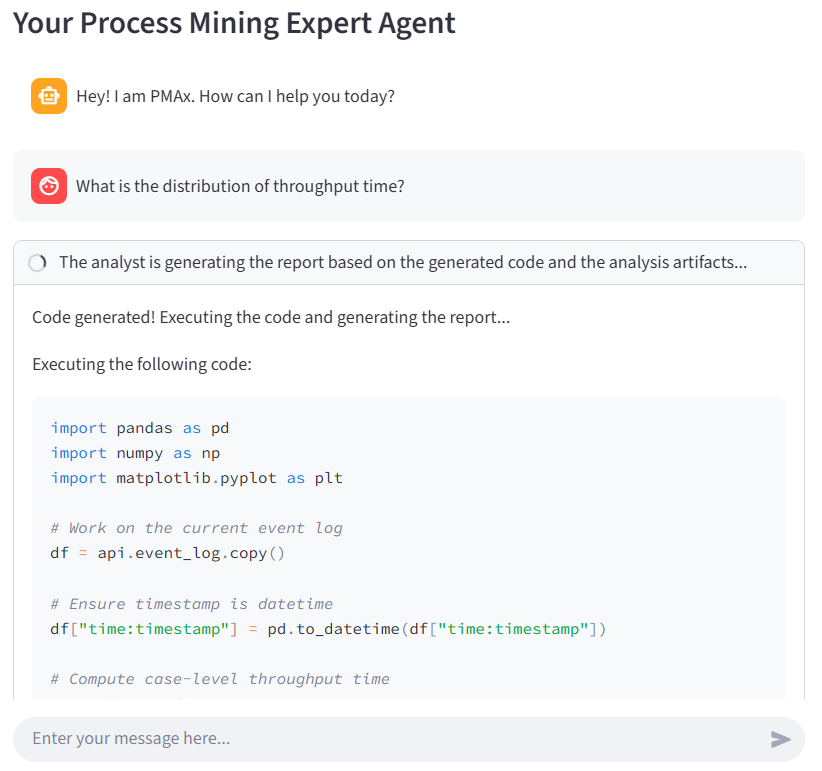}
    \caption{Code generation process.}
    \label{fig:codegen}
\end{subfigure}
\hfill
\begin{subfigure}[t]{0.48\linewidth}
    \centering
    \includegraphics[width=\linewidth]{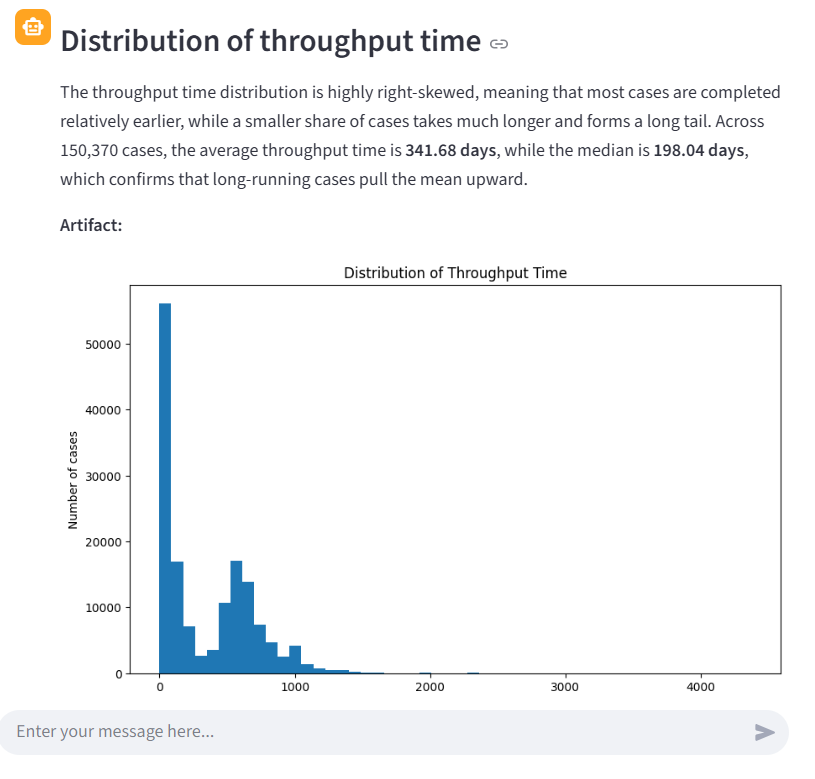}
    \caption{An excerpt from the generated analysis report.}
    \label{fig:report}
\end{subfigure}

\caption{PMAx conversational interface showing (a) real-time code synthesis and (b) the generated report.}
\label{fig:ui}
\end{figure}

PMAx is implemented as an extension within the open-source ProMoAI tool suite (\url{https://github.com/fit-process-mining/ProMoAI}). The workflow begins with a configuration step where the user selects their preferred AI provider, chooses a specific LLM model, and provides the necessary API key. Upon uploading an event log to start the session, the system transitions to a conversational interface (cf. \autoref{fig:ui}). To ensure complete transparency and auditability, the UI features a dedicated panel that allows users to monitor the real-time Python code synthesis performed by the Engineer Node (\autoref{fig:codegen}). Following the local execution of the generated code, the Analyst Node populates the view with a comprehensive, data-grounded report (\autoref{fig:report}), seamlessly combining narrative insights with the extracted visual evidence.


\section{Related Work}
The application of LLMs in process mining has evolved rapidly. Early efforts focused on translating textual descriptions into formal process models, such as BPMN or Petri nets \cite{DBLP:conf/models/IvanchikjSP20,DBLP:conf/bpmds/KouraniB0A24,safan2025bpmn}. Recent research has explored the direct reasoning capabilities of LLMs for a wider range of analytical tasks. For instance, Rebmann et al. \cite{DBLP:journals/procsci/RebmannSGA25} systematically evaluated the potential of LLMs to perform semantic tasks like activity prediction and anomaly detection by reasoning over textual process descriptions.

Building on the reasoning capabilities of LLMs, a significant research track has focused on creating conversational interfaces for process analysis. The primary goal of these systems is to translate a user's natural language question into a direct answer, often by converting the query into a formal language, e.g., SQL, or by having the LLM reason directly over abstracted data. Key examples include frameworks for enhanced process model comprehension through semantic querying \cite{DBLP:journals/dss/KouraniBHKWLAAS26}, conversational analysis of complex Object-Centric Event Logs (OCEL) to avoid manual query formulation \cite{Casciani2026}. While powerful, these systems typically act as ``translators'', providing answers without exposing the underlying data manipulation and code generation steps.

The domain is shifting toward autonomous agentic frameworks, a paradigm maturing in broader data science. Systems like AutoGen \cite{DBLP:journals/corr/abs-2308-08155} and LIDA \cite{DBLP:conf/acl/Dibia23} automate exploration and visualization, while self-correcting agents now autonomously write and debug code to solve complex analytical tasks \cite{DBLP:journals/corr/abs-2510-16872}.
This agentic approach is nascent in process mining. Berti et al. \cite{DBLP:journals/corr/abs-2408-07720} introduced a conceptual framework for re-thinking process mining tasks as multi-agent workflows. Furthermore, Vu et al. \cite{DBLP:conf/bpm/VuKLRK25} confirmed strong practitioner interest in agentic BPM but also highlighted significant governance challenges, such as the risk of LLM hallucinations and the need for reliable, verifiable outputs.

\section{System Architecture}
The system architecture is illustrated in \autoref{fig:overview}. The workflow begins when the user provides an event log, which undergoes a schema extraction process. This step produces an event log abstraction, ensuring that the agents receive a structural overview (e.g., column names, data types, and attribute samples) rather than the raw data. This approach addresses two critical requirements: it overcomes LLM context window constraints while simultaneously mitigating data privacy concerns by relying on data minimization. Concurrently, the user submits a natural language query, which is forwarded to the Engineer Node.

As a specialized code-generation agent, the Engineer Node synthesizes the necessary Python scripts. Before execution, the code is subjected to static analysis to verify adherence to pre-defined security and syntax constraints. Following execution, the Analyst Node receives the textual context and metadata of the generated artifacts. Based on this data-driven evidence, the Analyst Node generates a comprehensive report that addresses the user's initial question through a combination of natural language interpretation and visual evidence. We now describe each of the main components in detail.
\subsection{Collaborative Memory State}
To facilitate seamless collaboration between the Engineer and Analyst nodes, we implement a shared memory state, similar to the architecture described in \cite{DBLP:journals/corr/abs-2505-18279}, ensuring that pertinent information is synchronized across the different nodes. A core design philosophy of the method is the divide-and-conquer principle. We decompose the traditionally monolithic and complex task of process mining analysis into two specialized subtasks: (i) technical code synthesis and (ii) semantic result interpretation.

By isolating the conversation histories with each node within the state, we adhere to a strict separation of concerns. This ensures that the Analyst node is not influenced by code technicalities, which could otherwise exhaust the context window or lead to reasoning errors. Instead, the shared state acts as a filtered conduit, exchanging only the essential context and allowing each node to focus exclusively on its domain-specific expertise.

\subsection{Data Abstraction and Processing}
Real-world event logs frequently comprise thousands of traces and millions of events \cite{DBLP:books/sp/Aalst16}, presenting a significant challenge for LLMs due to current context window constraints \cite{DBLP:journals/corr/abs-2404-06035}. Additionally, many organizations do not wish to expose sensitive information. This is the main reason why agentic code generation cannot be applied directly to raw enterprise datasets. To mitigate these privacy and scalability concerns, we implement a metadata-driven abstraction layer that summarizes the log's structural schema and attribute types rather than providing the raw data. Furthermore, to ensure the agents are semantically grounded in process mining conventions, we provide specialized domain knowledge. This includes identifying the mandatory subset of attributes---namely case identifiers, timestamps, and activity names—and common naming conventions in the XES standard \cite{DBLP:journals/cim/AcamporaVSAGV17}.
\subsection{The Engineer Node}
The Engineer Node is a specialized LLM agent designed to bridge the gap between high-level natural language queries and the technical specificities of process mining algorithms. Its primary objective is to synthesize executable Python scripts that leverage a pre-defined toolkit of PM4Py and standard data science libraries.
\subsubsection{Context Injection}
To mitigate the risk of code hallucinations and ensure rigorous API alignment, the Engineer Node is initialized following the principle of role assignment \cite{DBLP:journals/corr/abs-2305-14688}, i.e., by explicitly defining the agent as a specialized Data Engineer. Beyond this role-based prompting, the framework injects a technical specification that establishes the operational environment and data-access protocols through four foundational preconditions:

\begin{enumerate}
\item \textbf{Data Access Layer:} The agent is informed that the target event log is accessible via a persistent Python object. This enables a code-to-data paradigm; rather than sending the full event log to the LLM, the agent generates code to be executed locally. This ensures the scalability of the method.
\item \textbf{Meta-Schema Exposure:} The \textit{Event Log Abstraction} is injected into the prompt, providing the agent with the precise column structure, data types, and attribute samples of the currently loaded log.
\item \textbf{Domain Knowledge:} The system identifies core process attributes (case IDs, activities, and timestamps) in a format-independent manner. This ensures compatibility with both XES and CSV files, providing structural context regardless of the file type.
\item \textbf{Operational Toolbox and API Constraints:} To ensure secure and valid code, an API object encapsulates core PM4Py functionalities \cite{DBLP:journals/simpa/BertiZS23} within a whitelisted environment. As detailed in \autoref{lst:api_specification}, the agent is restricted to standard data (Pandas \cite{reback2020pandas}, NumPy \cite{harris2020array}) and visualization libraries (Matplotlib \cite{Hunter:2007}, Plotly \cite{plotly}, Seaborn \cite{Waskom2021}). 
\end{enumerate}

By combining these four layers, the framework creates a constrained yet expressive environment where the agent can reason about process mining tasks without the common pitfalls of unconstrained LLM code generation.
\begin{lstlisting}[
    caption={API Specification and Operational Rules for the Engineer Agent.},
    label={lst:api_specification}, 
    frame=single, 
    float, 
    floatplacement='!t', 
    basicstyle=\scriptsize\ttfamily,
    breaklines=true,
    columns=fullflexible,
    emph={api, filter_time_range, filter_attribute, filter_pandas_query, get_dfg_summary, get_model_summary, get_variant_summary, get_case_summary, discover_process_model, cc_alignments, cc_token_based_replay, save_pnet, save_visualization, save_dataframe},
    emphstyle=\color{blue}\bfseries
]
[CATEGORY: FILTERING]
- api.filter_time_range(start: str, end: str)   # Temporal subsetting
- api.filter_attribute(column: str, value: str) # Attribute-based filtering
- api.filter_pandas_query(query: str)           # Complex logic (e.g., "amount > 500")

[CATEGORY: ABSTRACTION & SUMMARIZATION]
- api.get_dfg_summary()      # Markovian/DFG abstraction
- api.get_model_summary()    # Petri net/Process model abstraction
- api.get_variant_summary()  # Unique sequence analysis
- api.get_case_summary()     # Pattern and outlier detection

[CATEGORY: MINING & CONFORMANCE]
- api.discover_process_model()  # Updates state with discovered Petri net
- api.cc_alignments()           # Returns (fitness, precision, F1)
- api.cc_token_based_replay()   # Returns (fitness, precision, F1)

[CATEGORY: VISUALIZATION & PERSISTENCE]
- api.save_pnet()                             # Exports Petri net
- api.save_visualization(fig, desc, data)     # Persists Matplotlib/Plotly/Seaborn
- api.save_dataframe(df, desc)                # Persists results/intermediate tables

[OPERATIONAL RULES]
1. LIBRARY RESTRICTION: Use only whitelisted libraries (pm4py, pandas, numpy, plotly).
2. PERSISTENCE: Always use api.save_visualization/dataframe; built-in I/O is disabled.
3. EFFICIENCY: Generate visualizations sparsely to minimize computational overhead.
4. SELF-CONTAINED: Include all necessary import statements for non-API libraries.
\end{lstlisting}
\subsubsection{Code Generation}
Synthesized code is statically verified for library compliance and security, restricting I/O to the framework's API to ensure system integrity. Validated scripts execute in-memory; any runtime exception triggers a self-correction loop where the captured traceback guides iterative refinement until successful execution or a maximal iteration threshold is reached.

\subsection{The Analyst Node}
\subsubsection{Report Synthesis}
The Analyst Node translates technical artifacts into insights while remaining isolated from the code. It employs an adaptive persistence protocol to stay within context limits: small dataframes are serialized as Markdown, while larger ones are summarized via statistics. To optimize tokens, the framework transmits raw data sources instead of high-overhead images for visualization interpretation. Multi-turn interactions are handled through incremental updates, transferring only new artifacts per iteration. Finally, the agent reconciles this evidence with the user query to produce a domain-specific report
\vspace{-10pt}
\subsubsection{Error Handling}
To mitigate hallucinations, a strict output schema—a list of typed dictionaries (text or artifact)—enforces deterministic artifact referencing. Structural non-conformance or invalid references trigger an automated error-handling cycle, ensuring the report is strictly grounded in the execution results. The validation error is fed back to the LLM, triggering a self-correction loop that forces the agent to reconcile its report with the empirical evidence stored in the shared state.

\section{Tool Demonstration: Loan Application Process Analysis}
To demonstrate the practical utility of PMAx, we used the framework to analyze the BPI Challenge 2017 dataset \cite{BPIC2017}, a real-world loan application log from a Dutch financial institution. We framed our investigation around the business-driven questions addressed by \cite{badakhshan2017bpi}, focusing on: \textbf{(Q1)} typical workflow discovery, \textbf{(Q2)} throughput time distribution, \textbf{(Q3)} waiting time and bottleneck identification, \textbf{(Q4)} the impact of information requests on offer acceptance, and \textbf{(Q5)} success rates for single vs. multiple offers.

The report was generated using OpenAI's GPT-5.4 \cite{DBLP:journals/corr/abs-2601-03267}. In all scenarios, the Engineer Node successfully synthesized executable Python code, which we manually verified for technical correctness. While the Analyst Node's reports were occasionally verbose, they provided accurate data-grounded insights that directly addressed the imposed questions. Notably, complex comparative tasks (e.g., Q5) were resolved autonomously, confirming that PMAx can transform high-level business queries into precise, domain-specific code. The generated artifacts and report are available online.\footnote{\url{https://github.com/fit-process-mining/PMAx-evaluation}}
\section{Conclusion}
In this paper, we presented PMAx, a framework that shifts the focus of process mining from algorithmic development to agentic system design. By implementing a closed-loop orchestration specifically tailored to bridge the gap between process semantics and technical implementation, the framework effectively decouples technical code synthesis from semantic interpretation, thereby overcoming the limitations of LLM hallucinations, constrained context windows, and privacy concerns. Our tool demonstration using a real-world dataset shows that the system autonomously generates data-grounded insights from natural language cues, marking a significant step toward democratizing process mining through ``no-code'' analytical interfaces.

Future work will focus on extending this paradigm to Object-Centric Process Mining (OCPM) \cite{math11122691}. We envision a transition where agentic workflows are utilized to manage OC event data, translating user questions into sophisticated relational or graph-based queries to handle the increasing complexity of modern organizational processes. Additionally, we intend to move beyond linear workflows toward autonomous inter-agent dialogue, enabling the framework to dynamically resolve analytical ambiguities while further compressing the shared state context

%
%
%
\bibliographystyle{splncs04}
\bibliography{mybibliography}

\end{document}